\documentclass[letterpaper, 10 pt, conference]{ieeeconf} 

\IEEEoverridecommandlockouts 

\usepackage{cite}
\usepackage{graphicx} 
\graphicspath{{./figures/}}
\usepackage{epsfig} 
\usepackage{mathptmx} 
\usepackage{times} 
\usepackage{amsmath} 
\usepackage{amssymb} 
\usepackage[skip=0.3ex]{subcaption}
\usepackage{placeins}
\usepackage{float}
\usepackage{xcolor}
\usepackage{bm}

\DeclareMathAlphabet      {\mbi}{OML}{cmm}{b}{it}
\DeclareMathAlphabet      {\mathit}{OML}{cmm}{}{it}
\DeclareMathAlphabet{\mathcal}{OMS}{cmsy}{m}{n}

\DeclareMathOperator{\q}{\mbi{q}}
\DeclareMathOperator{\dq}{\dot{\mbi{q}}}
\DeclareMathOperator{\Jq}{\mbi{J(q)}}

\DeclareMathOperator{\Jb}{\mbi{J}_b\mbi{(q)}}

\DeclareMathOperator{\Jbar}{\mbi{\Bar{J}(q)}}
\DeclareMathOperator{\Lb}{\mbi{\Lambda}}

\DeclareMathOperator{\g}{\mbi{G(q)}}

\DeclareMathOperator{\pex}{\mbi{p}_{Ex}}
\DeclareMathOperator{\Rex}{\mbi{R}_{Ex}}
\DeclareMathOperator{\Hex}{\mbi{g}_{Ex}}
\DeclareMathOperator{\peb}{\mbi{p}_{Eb}}
\DeclareMathOperator{\Reb}{\mbi{R}_{Eb}}
\DeclareMathOperator{\Heb}{\mbi{g}_{Eb}}

\DeclareMathOperator{\expE}{{\text{exp}}_{SE(3)}}

\newcommand\numberthis{\addtocounter{equation}{1}\tag{\theequation}}

\begin{document}

\title{\LARGE \bf Whole-Body Bilateral Teleoperation of a Redundant Aerial Manipulator}

\author{Andre Coelho, Harsimran Singh, Konstantin Kondak, Christian Ott
\thanks{The authors are with the Institute of Robotics and Mechatronics of the
German Aerospace Center (DLR), Oberpfaffenhofen, Germany}%
\
\thanks{{\tt\small Andre.Coelho@dlr.de}}}

%
%

%


\maketitle

\begin{abstract}
Attaching a robotic manipulator to a flying base allows for significant improvements in the reachability and versatility of manipulation tasks. In order to explore such systems while taking advantage of human capabilities in terms of perception and cognition, bilateral teleoperation arises as a reasonable solution. However, since most telemanipulation tasks require visual feedback in addition to the haptic one, real-time (task-dependent) positioning of a video camera, which is usually attached to the flying base, becomes an additional objective to be fulfilled. Since the flying base is part of the kinematic structure of the robot, if proper care is not taken, moving the video camera could undesirably disturb the end-effector motion. For that reason, the necessity of controlling the base position in the null space of the manipulation task arises. In order to provide the operator with meaningful information about the limits of the allowed motions in the null space, this paper presents a novel haptic concept called Null-Space Wall. In addition, a framework to allow stable bilateral teleoperation of both tasks is presented. Numerical simulation data confirm that the proposed framework is able to keep the system passive while allowing the operator to perform time-delayed telemanipulation and command the base to a task-dependent optimal pose. 
\end{abstract}

\IEEEpeerreviewmaketitle

\setlength{\textfloatsep}{10pt}
\section{Introduction}
The use of unmanned aerial vehicles (UAV) as a flying base for robotic manipulators has been object of intensive research in the recent years \cite{ollero18,ruggiero18,khamseh18,orsag18}. One of the main goals of such systems is to replace and assist humans in tasks as inspection and repairing of bridges, high-voltage electric lines, and wind-turbine blades \cite{ollero18,ruggiero18}. The use of such systems help overcome a series of limitations of fixed-based robots, e.g., limited workspace. Some recent works focused on performing aerial manipulation tasks autonomously \cite{thomas13,kondak14,jimenez19}. However, in order to more deeply explore the manipulation capacity of such systems, telemanipulation arises as an alternative \cite{gioioso14, mohammadi16,islam19,lee20}. Adding a human operator to the loop allows for the simplification of many of the system features, especially in what concerns perception and reasoning.
\par
\begin{figure}[tb]
\centering
\includegraphics[trim={0.04cm 0cm 0cm 0cm},clip,width=0.9\linewidth]{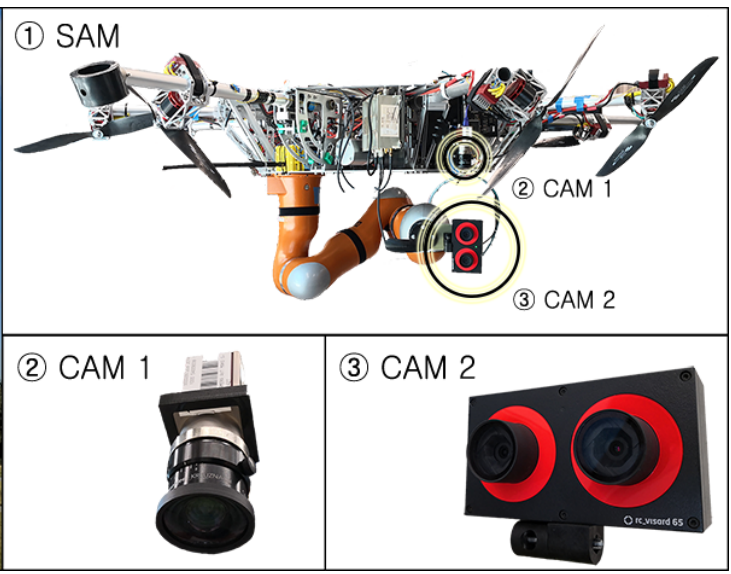}
\caption{(1) DLR Suspended Aerial Manipulator (SAM). (2)~eye-to-hand camera (CAM 1). (3) hand-eye camera (CAM 2).}
\label{fig:sam}
\end{figure}

Nevertheless, during field experiments performed in the scope of the AEROARMS project \cite{ollero18,lee20}, an additional condition for aerial telemanipulation has been found. Since tasks are usually performed without direct eye contact, the use of video cameras to stream images to the distant operator is a necessary add-on to haptic feedback. Adding to that, in order to facilitate the operator's reasoning, a first-person or eye-to-hand view, i.e., a view of the entire arm, has been found beneficial. In the case of aerial manipulators, like the DLR Suspended Aerial Manipulator (SAM) \cite{sarkisov2019,sarkisov20,lee20}, a camera is usually attached to the flying base in order to provide an eye-to-hand view (see Fig.~\ref{fig:sam}). If correctly positioned, the camera image allows the operator to have a view of the arm and the area of interest for manipulation. 
\par
In case the aerial manipulator is redundant, there might be some freedom to move the base and, consequently, the video camera without affecting the end-effector. For such systems, defining the camera pose autonomously could allow for a better view to the operator. However, since its optimal location is rather subjective and task-dependent, it would be reasonable to provide the operator with some authority to control that task without disturbing the end-effector pose. Adding to that, since the set of possible base poses for a given end-effector pose is usually a smaller subset of the Cartesian-space, the master device can be used in order to inform the operator if the pose to which the base is being commanded is approaching the limits of that subset.
\par
In light of that, this paper proposes a framework to allow a human operator to teleoperate the flying base of a redundant aerial manipulator in order to achieve a desired view of the task being performed while receiving meaningful haptic feedback about the reachability of the commanded pose without disturbing the end-effector. For that purpose, a previously presented whole-body control framework is applied in order to define a task hierarchy, where the positioning of the base is performed in the null space \cite{murray94} of the main telemanipulation task. Adding to that, a novel haptic concept called \textit{Null-Space Wall} is introduced in order to inform the operator in an intuitive way whether the pose to where the base is being commanded is reachable. This concept is defined in such a way that no additional fixtures have to be added in order to create haptic feedback. Elseways, the already computed forces from the whole-body controller are explored. The use of haptic feedback in order to constrain the commanded motion has been extensively explored in the literature (e.g., \cite{boessenkool12,masone18,marinho19}). However, to the best of the authors' knowledge this is the first time that such a technique is applied in order to allow hierarchical whole-body telemanipulation of redundant robots.
\par
In order to evaluate the proposed framework, bilateral teleoperation of the SAM (see Fig.~\ref{fig:sam}) is simulated with round-trip time-delays up to 300 ms. The stability and efficacy of the approach in providing the user with meaningful haptic information about the limits of the null space can be verified in the data.


\section{Time Domain Passivity Approach}
\label{sec:tdpa}
Time Domain Passivity Approach (TDPA) \cite{hannaford02,ryu10,coelho18} was developed in order to enforce stability of both haptic and bilateral teleoperation setups where velocity and force signals are exchanged. In teleoperation systems, time-delay and package losses introduced by the channel might compromise the overall stability of the system, whereas in haptics instability can be a result of sampling.  \par
\par
\begin{figure}[tbhp]
\centering
\includegraphics[trim={7.4cm 8cm 9.6cm 7.4cm},clip,width=0.8\linewidth]{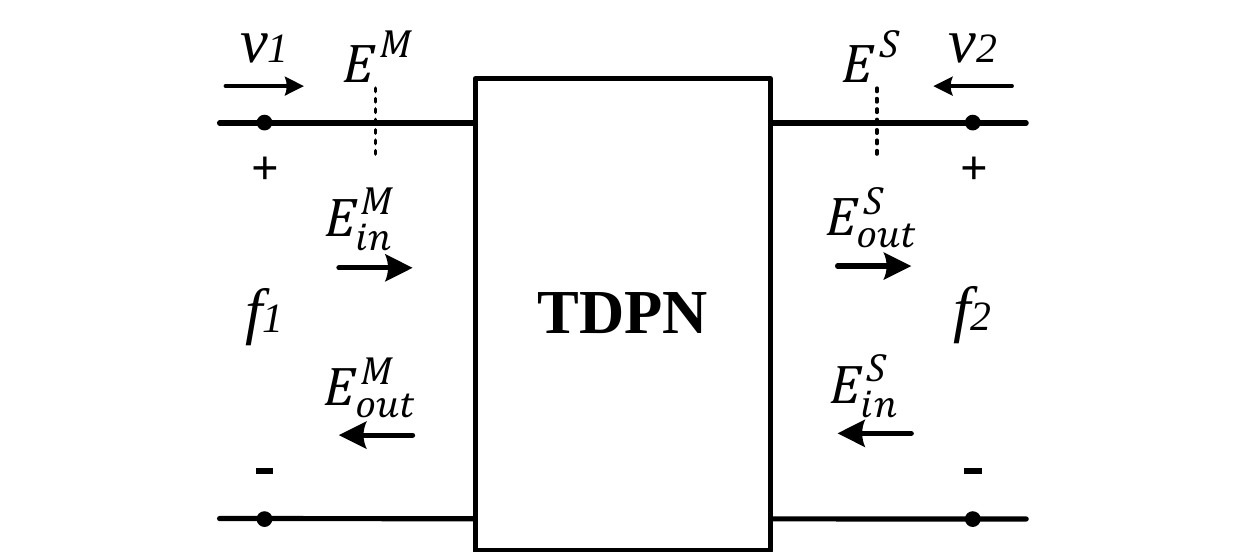}
\caption{Signal flow of the TDPN.}
\label{fig:tdpn_flow}
\end{figure}

In bilateral teleoperation setups, the communication channel is usually represented by one or more Time Delay Power Networks (TDPNs, \cite{ryu10}), which are two-port networks that exchange velocities and forces. In addition to constant or variable time-delays, TDPNs can also model package losses in the signals being transmitted. Fig.~\ref{fig:tdpn_flow} shows the signal flow of the TDPN. $E^M$ and $E^S$ are the energies computed on the master and slave sides, respectively. The $in$ and $out$ subscripts are used to represent the direction of flow, namely into or out of the channel. 
\par
 The pairs $v_1$/$f_1$ and $v_2$/$f_2$ from Fig.~\ref{fig:tdpn_flow} are the flow-effort pairs on each side of the TDPN, such that in coordinates
 \begin{align} \label{eq:energy_integrals}
E^M(k)  = \Delta T \sum_{j=1}^{k} f_1(j)^{\mathit{T}} v_1(j) \, , \\
E^S(k) = \Delta T \sum_{j=1}^{k} -f_2(j)^{\mathit{T}} v_2(j) \, ,
\end{align}
where $\Delta T$ is the sampling time and $k$ is the discrete-time index.
\par
A sufficient condition for passivity of a TDPN is that
\begin{align} 
E^{L2R}_{obs}(k) = E^M_{in}(k-T_f(k))-E^S_{out}(k) \geq 0, \quad \forall k \geq 0, \label{eq:energy_obs} \\
E^{R2L}_{obs}(k) = E^S_{in}(k-T_b(k))-E^M_{out}(k) \geq 0, \quad  \forall k \geq 0, \label{eq:energy_obs1}
\end{align} 
where $E^{L2R}_{obs}(k)$ and $E^{R2L}_{obs}(k)$ are the observed left-to-right and right-to-left energy flows observed on the right and left-hand sides of the TDPN. $T_f(k)$ and $T_b(k)$ are the forward and backward delays, respectively.
\par
\begin{figure}[tbhp]
\centering
\includegraphics[trim={6cm 6.7cm 4.8cm 6cm},clip,width=1\linewidth]{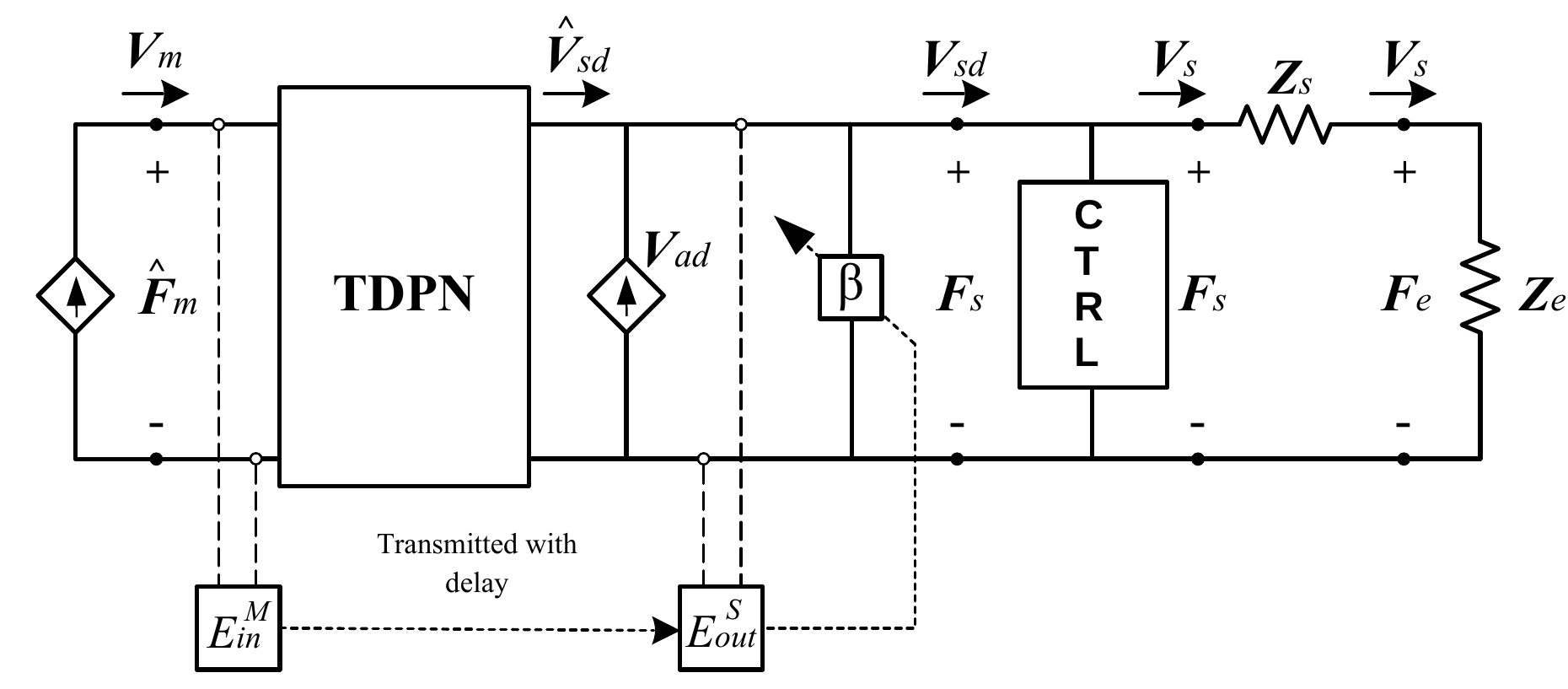}
\caption{Slave side of a P-F architecture. The passivity observer-passivity controller (PO-PC) pair  is applied in admittance configuration ($\mbi{\beta}$). $\mbi{V}_{ad}$ is the drift-compensation velocity.}
\label{fig:drift_comp}
\end{figure}

One of the most common teleoperation schemes is the P-F architecture \cite{lawrence93}, where the master velocity is sent through the channel and serves as desired velocity to the slave. In turn, the force produced by the slave-side controller is sent back to the master. Following the framework presented by Artigas et al. \cite{artigas11}, using a hybrid of circuit and network representation, the slave side of the P-F architecture can be represented as shown in Fig.~\ref{fig:drift_comp}. There, the communication channel is represented by a TDPN; $\mbi{V}_m$ and $\mbi{V}_s$ are the velocities of the master and slave devices; $\mbi{F}_s$ is the force exerted by the slave-side controller; $\mbi{V}_{sd}$ is the delayed master velocity; and $\hat{\mbi{F}}_m$ is the delayed slave control force applied to the master device. Adding to that, $\mbi{V}_{ad}$ and $\mbi{\beta}$ are the the drift compensation velocity source \cite{coelho18,coelho19} and the admittance-type passivity controller, which will be addressed subsequently.

\subsubsection{Passivity Observer}
In order to take into account the energy removed by the passivity controllers up to the previous time steps ($E^M_{PC}(k-1)$ and $E^S_{PC}(k-1)$), the energy flow on each side of the TDPN is computed as 
\begin{align} \label{eq:po}
W_M(k) = E^S_{in}(k-T_b(k))-E^M_{out}(k)+E^M_{PC}(k-1), \\
W_S(k) = E^M_{in}(k-T_f(k))-E^S_{out}(k)+E^S_{PC}(k-1). 
\label{eq:po2}
\end{align} 
\subsubsection{Passivity Controller}
\label{sec:pc}
The passivity controller (PC) acts as an adaptive damping in order to guarantee the passivity of the channel. It can be applied in impedance or admittance configuration, according to the architecture requirements. In Fig.~\ref{fig:drift_comp} the PC ($\mbi{\beta}$) is being applied in admittance configuration in order to modify the velocity coming out of the channel. 

\par
In order to fulfill the passivity conditions from (\ref{eq:energy_obs}) and (\ref{eq:energy_obs1}) in a P-F architecture, the passivity controller is applied in admittance configuration on the slave side and in impedance configuration on the master side.
\par
The admittance-type PC can be applied as \cite{coelho19}
\begin{equation}
    \mbi{\beta}(k)=d_f(k)\mbi{\Gamma}(k) \, ,
\end{equation}
\begin{equation} \label{eq:beta}
d_f(k)=
\begin{cases}
    0                                                        & \text{if } W_S(k)>0\\
    -\cfrac{ W_S(k)}{\Delta T||\mbi{F}_{s}(k)||^2_{\Gamma}} & \text{else, if } ||\mbi{F}_{s}(k)||^2_\Gamma>0 \,,
\end{cases}
\end{equation}
where
\begin{equation}
    ||\mbi{F}_{s}(k)||^2_\Gamma=\mbi{F}_{s}(k)^\mathit{T}\mbi{\Gamma}(k)\mbi{F}_{s}(k) \, ,
\end{equation}
and $\mbi{\Gamma}(k)$ is a symmetric positive-definite weighting matrix.
\par
The velocity removed by the PC from the delayed master velocity in order to keep the system passive will be
\begin{equation}
    \mbi{V}_{pc}(k)=\mbi{\beta}(k)\mbi{F}_{s}(k) \, ,
\end{equation}
and the resulting velocity used as a reference by the slave will be
\begin{equation} \label{eq:pc_imp}
\mbi{V}_{sd}(k)=\hat{\mbi{V}}_{sd}(k)- \mbi{V}_{pc}(k)\, ,
\end{equation}
assuming all velocities are represented in the same frame. 
\par
Likewise, the impedance-type passivity controller can be applied as \cite{ott11}
\begin{equation}
    \mbi{\alpha}(k)=d_v(k)\mbi{\Psi}(k) \, ,
\end{equation}
\begin{equation} \label{eq:alpha}
d_v(k)=
\begin{cases}
    0                                                        & \text{if } W_M(k)>0\\
    -\cfrac{ W_M(k)}{\Delta T||\mbi{V}_{m}(k)||^2_{\Psi}} & \text{else, if } ||\mbi{V}_{m}(k)||^2_\Psi>0 \,,
\end{cases}
\end{equation}
where
\begin{equation}
    ||\mbi{V}_{m}(k)||^2_\Psi=\mbi{V}_{m}(k)^\mathit{T}\mbi{\Psi}(k)\mbi{V}_{m}(k) \, ,
\end{equation}
and $\mbi{\Psi}(k)$ is a symmetric positive-definite weighting matrix.
\par
The force removed by the passivity controller from the delayed slave force in order to keep the system passive will be
\begin{equation}
    \mbi{F}_{pc}(k)=\mbi{\alpha}(k)\mbi{V}_{m}(k) \, ,
\end{equation}
and the resulting velocity used as a reference by the slave will be
\begin{equation} \label{eq:pc_adm}
\mbi{F}_{m}(k)=\hat{\mbi{F}}_{m}(k)- \mbi{F}_{pc}(k)\, .
\end{equation}

\section{Whole-Body Bilateral Teleoperation of Redundant Aerial Manipulators}
\subsection{Notations and Definitions}
\label{sec:notation}
\subsubsection{The Special Euclidean group and its Lie algebra} The pose of a rigid body in space can be represented by elements of the special Euclidean group $SE(3)$, whose matrix form is
\begin{equation}
\mbi{g}=\begin{bmatrix}\mbi{R} &  \mbi{p} \\\mathbf{0} & 1\\\end{bmatrix} \in \: SE(3) \, ,
\end{equation}
where $\mbi{p} \in \mathbb{R}^3$ describes the position vector and $\mbi{R}$ is an element of the Special Orthogonal group $SO(3)$ relative to the rotation of the body. Elements of $SE(3)$ can also be identified as $\mbi{g}=(\mbi{R},\,\mbi{p})$. Furthermore, the velocity of a rigid body can be expressed by elements of the Lie algebra of $SE(3)$, namely $\mathfrak{se}(3)$, as
\begin{equation}
[\mbi{V}]^\wedge=\begin{bmatrix}\widehat{\mbi{\omega}} &  \mbi{v} \\\mathbf{0} & 0\\\end{bmatrix} \: \in \mathfrak{se}(3)\,, 
\end{equation}
where $ \widehat{\cdot}$ indicates the skew-symmetric operator applied to a vector in $\mathbb{R}^3$ and $\mbi{\omega}, \, \mbi{v} \in \mathbb{R}^3$ are angular and linear velocities, respectively. Adding to that, due to the isomorphism between $\mathfrak{se}(3)$ and $\mathbb{R}^6$, it is useful to define the operator $[\cdot]^\wedge \, : \, \mathbb{R}^6 \rightarrow \mathfrak{se}(3)$, such that the velocities of rigid bodies can be expressed as elements of $\mathbb{R}^6$ or $\mathfrak{se}(3)$. Such velocities can be represented in body (${}^B \mbi{V}$), or in spatial frame (${}^S \mbi{V}$) \cite{murray94}. 
\subsubsection{Dynamical systems in $SE(3)$}
A dynamical system with state $\mbi{g} \in SE(3)$ and body velocity $[{}^B\mbi{V}]^\wedge \in \mathfrak{se}(3)$ evolves according to the following differential equation in continuous time \cite{bullo95}
\begin{equation}
\label{eq:g_dot}
    \dot{\mbi{g}}(t)=\mbi{g}(t)\,[{}^B\mbi{V}(t)]^\wedge\, ,
\end{equation}
whose recursive solution in discrete time, given a set of initial conditions, can be approximated to
\begin{align}
    \mbi{g}(k)&=\mbi{g}(k-1)\expE\left([{}^B\mbi{V}(k)]^{\wedge}\Delta T\right) \, ,
\label{eq:int_body}
\end{align}
where $\expE$ is the $SE(3)$ exponential map \cite{bullo95,coelho19}.
\subsection{Decoupling Control of Redundant Aerial Manipulators}
\label{sec:ctrl}
The dynamics of a fully actuated aerial manipulator with $n$ joints can be written as
\begin{equation}
   \mbi{ M(q)\Ddot{q}+C(q,\dot{q})\dot{q}+G(q)}=\mbi{\tau} + \mbi{\tau}_{ext} \, ,
   \label{eq:dyn}
\end{equation}
where $\mbi{q} \in \mathbb{R}^n$ is a set of generalized coordinates, $\mbi{M(q)} \in \mathbb{R}^{n \times n}$ is the inertia matrix, $\mbi{C(q,\dot{q})\dot{q}} \in \mathbb{R}^n$ is a vector of Coriolis and centrifugal forces, and $\mbi{G(q)} \in \mathbb{R}^n$ is the gravitational generalized torque vector. $\mbi{\tau} \in \mathbb{R}^n$ and $\mbi{\tau}_{ext} \in \mathbb{R}^n$ are the generalized control and external torque vectors, respectively. 
\par
Following the framework presented by Ott et al. \cite{ott15} for hierarchical whole-body control of kinematically redundant robots, where the number of minimal end-effector pose coordinates, $\mbi{\xi(q)} \in \mathbb{R}^m$, is less than the number of joint generalized coordinates, i.e., $m<n$, the set of task coordinates can be extended to
\begin{equation}
    \begin{bmatrix}
    \mbi{V_x} \\
    \mbi{V_n} \\
    \end{bmatrix}
    =
    \mbi{\Bar{J}(q)\dot{q}}=
    \begin{bmatrix}
    \mbi{J(q)} \\
    \mbi{N(q)} \\
    \end{bmatrix} \mbi{\dot{q}} \, ,
\end{equation}
where $\mbi{V_x} \in \mathbb{R}^{m}$ and $\mbi{V_n} \in \mathbb{R}^{n-m}$ are the set of Cartesian and null space velocities, respectively. $\mbi{J(q)}$ is the Jacobian matrix that maps generalized to end-effector velocities. In this paper, it is assumed that $\mbi{V_x} \in \mathbb{R}^6$ is the body velocity of the end-effector with respect to some fixed inertial frame. Therefore $\mbi{J(q)}$ is the \textit{body Jacobian} \cite{murray94} of the Cartesian task. Under the assumption of full row rank of $\mbi{J(q)}$, the matrix $\mbi{N(q)}$ can be constructed as \cite{ott15}
\begin{equation}
\label{eq:N}
    \mbi{N(q)=(Z(q)M(q)Z(q)}^\mathit{T})^{-1}\mbi{Z(q)M(q)} \, ,
\end{equation}
where $\mbi{Z(q)}$ is a full row rank null space base matrix, such that $\mbi{J(q)Z(q)}^\mathit{T}=\mathbf{0}$. 
\par
Such formulation allows the manipulator dynamics to be written as 
\begin{align*}
    & \begin{bmatrix}\mbi{\Lambda_x(q)} & 0 \\ 0 & \mbi{\Lambda_n(q)} \\ \end{bmatrix} \begin{bmatrix}\mbi{\dot{V}_x} \\ \mbi{\dot{V}_n} \\\end{bmatrix} + \begin{bmatrix}\mbi{\mu_x(q,\dot{q})} & \mbi{\mu_{xn}(q,\dot{q})} \\ \mbi{\mu_{nx}(q,\dot{q})} & \mbi{\mu_n(q,\dot{q})} \\ \end{bmatrix}\begin{bmatrix}\mbi{V_x} \\ \mbi{V_n} \\\end{bmatrix} \\ &= \Jbar^\mathit{-T}(\mbi{\tau}+\mbi{\tau}_{ext}-\mbi{G(q))}\, .  \numberthis
\end{align*}
where $\mbi{\Lambda_x(q)}$ and $\mbi{\Lambda_n(q)}$ are the elements of a block-diagonal Cartesian inertia matrix $\Lb(\q)$. In addition, $\mbi{\mu_x}(\q,\dq)$, $\mbi{\mu_{xn}}(\q,\dq)$, $\mbi{\mu_{nx}}(\q,\dq)$, and $\mbi{\mu_n}(\q,\dq)$ are the elements of the Cartesian Coriolis-centrifugal matrix $\mu(\q,\dq)$.
\par
In order to control the pose of the flying base as the secondary task and the end-effector as the primary one, the following hierarchically-decoupling control law can be applied \cite{ott15}
\begin{align}
\label{eq:ctrl1}
    \mbi{\tau}&=\mbi{\tau}_x+\mbi{\tau}_n+\mbi{\tau}_\mu + \g \, , \\
    \mbi{\tau}_x &= \Jq^{\mathit{T}} \mbi{F_x} \, ,  \label{eq:ctrl1_2} \\ 
    \mbi{\tau}_n &= \mbi{N(q)}^{\mathit{T}} \mbi{Z(q)} \Jb^{\mathit{T}} \mbi{F}_b \, , \label{eq:ctrl1_3}  \\
    \mbi{\tau}_\mu &= \Jbar^{\mathit{T}}  \begin{bmatrix} \mathbf{0} & \mbi{\mu}_{xn}(\q,\dq) \\ \mbi{\mu}_{nx}(\q,\dq) & \mathbf{0} \\ \end{bmatrix}\begin{bmatrix}\mbi{V_x} \\ \mbi{V_n} \\\end{bmatrix} \, ,
    \label{eq:ctrl2}
\end{align}
where $\Jb$ is the Jacobian that maps the joint velocities $\dq$ to the body velocities of the flying base $\mbi{V}_b \in \mathbb{R}^6$. Moreover, $\mbi{F_x}$ is a Cartesian-space wrench applied to the end-effector and $\mbi{F}_b$ is the control wrench of the secondary task, before being projected into the null space of the primary one. Both wrenches are assumed to be body wrenches \cite{murray94}.
\par
In order to control the pose of the end-effector and of the flying base (as a secondary task), the following error elements can be defined
\begin{equation}
    \Hex =   \mbi{g}_{x,des}^{-1} \, \mbi{g}_{x} = (\Rex,\,\pex) \: , \: \Heb =  \mbi{g}_{b,des}^{-1} \, \mbi{g}_{b} = (\Reb,\,\peb) \, ,
\end{equation}
where $\mbi{g}_{x,des}$, $\mbi{g}_{x}$, $\mbi{g}_{b,des}$, and $\mbi{g}_{b}$ are homogeneous transformation matrices in $SE(3)$, which describe the desired and current poses of the end-effector (subscript $x$) and of the flying base (subscript $b$) with respect to a fixed inertial frame. The body velocities relative to these elements are $\mbi{V}_{x,des}$, $\mbi{V}_{x}$, $\mbi{V}_{b,des}$, and $\mbi{V}_{b}$, respectively, which relate to their time derivative according to (\ref{eq:g_dot}).
\par
To accomplish the desired tasks, the body Cartesian control wrenches $\mbi{F_x}$ and $\mbi{F}_b$ can be defined as \cite{zhang2000,ott11b}
\begin{align}
    \mbi{F_x}&=\begin{bmatrix} 
    -\Rex^{\mathit{T}} \mbi{K}_{Px} \pex
    \\
    -2 \Rex^{\mathit{T}} \mbi{E}(\eta_{Ex},\epsilon_{Ex})^{\mathit{T}}\mbi{K}_{Ox} \epsilon_{Ex}
    \end{bmatrix} + \mbi{K}_{Dx}\mbi{V}_{Ex} \, , \\
    \mbi{F}_b&=\begin{bmatrix} 
    -\Reb^{\mathit{T}} \mbi{K}_{Pb} \peb    \\
    -2 \Reb^{\mathit{T}} \mbi{E}(\eta_{Eb},\epsilon_{Eb})^{\mathit{T}}\mbi{K}_{Ob} \epsilon_{Eb}
    \end{bmatrix} + \mbi{K}_{Db}\mbi{V}_{Eb} \, ,
    \label{eq:Fb}
\end{align}
where the matrices $\mbi{K}_{(.)}$ are positive definite gain matrices. Furthermore, $\eta_{Ex}$ and $\epsilon_{Ex}$ are the scalar and vector parts of a quaternion representation of $\Rex$, respectively, and $\mbi{E}(\eta_{Ex},\epsilon_{Ex}) = \eta_{Ex}\mbi{I}_3 - \widehat{\epsilon}_{Ex}$. Moreover, $\mbi{V}_{Ex}$ and $\mbi{V}_{Eb}$ are the body velocities of $\mbi{g}_{Ex}$ and $\mbi{g}_{Eb}$, respectively.
\par
It is important to note in (\ref{eq:ctrl1_3}) that the control force $\mbi{F}_b$ goes through a projection before being commanded to the actuators. In this projection, the components of $\mbi{F}_b$ that conflict with the primary task are projected into zero and will not be commanded to the actuators via $\mbi{\tau_n}$.

\subsection{Null-Space Wall Concept}
Although $\mbi{F}_b$ is not always entirely applied to the base, it carries out important information about the limits of the null space of the primary task. In case the base is teleoperated, the operator directly commands $\mbi{g}_{b,des}$ and $\mbi{V}_{b,des}$. However, in case a position which is not reachable without affecting the end-effector position is commanded, the resulting force $\mbi{F}_b$ will be projected to zero. If the operator keeps commanding $\mbi{V}_{b,des}$ towards that direction, the errors between desired and current pose in (\ref{eq:Fb}) will increase. As a consequence, an increase in $\mbi{F}_b$ will be observed. This behavior resembles what happens when the robot being teleoperated hits a wall. Therefore, by sending $\mbi{F}_b$ as haptic feedback, the operator will have the feeling as if the robot were trying to penetrate a wall when an unfeasible pose is commanded to the flying base. The stiffness of that wall will be initially defined by $\mbi{K}_{Pb}$ and $\mbi{K}_{Ob}$, but could be scaled according to the task requirements. 
\par
\begin{figure}[t]
 \centering
 \includegraphics[trim={6cm 5cm 10cm 4.5cm},clip,width=0.8\linewidth]{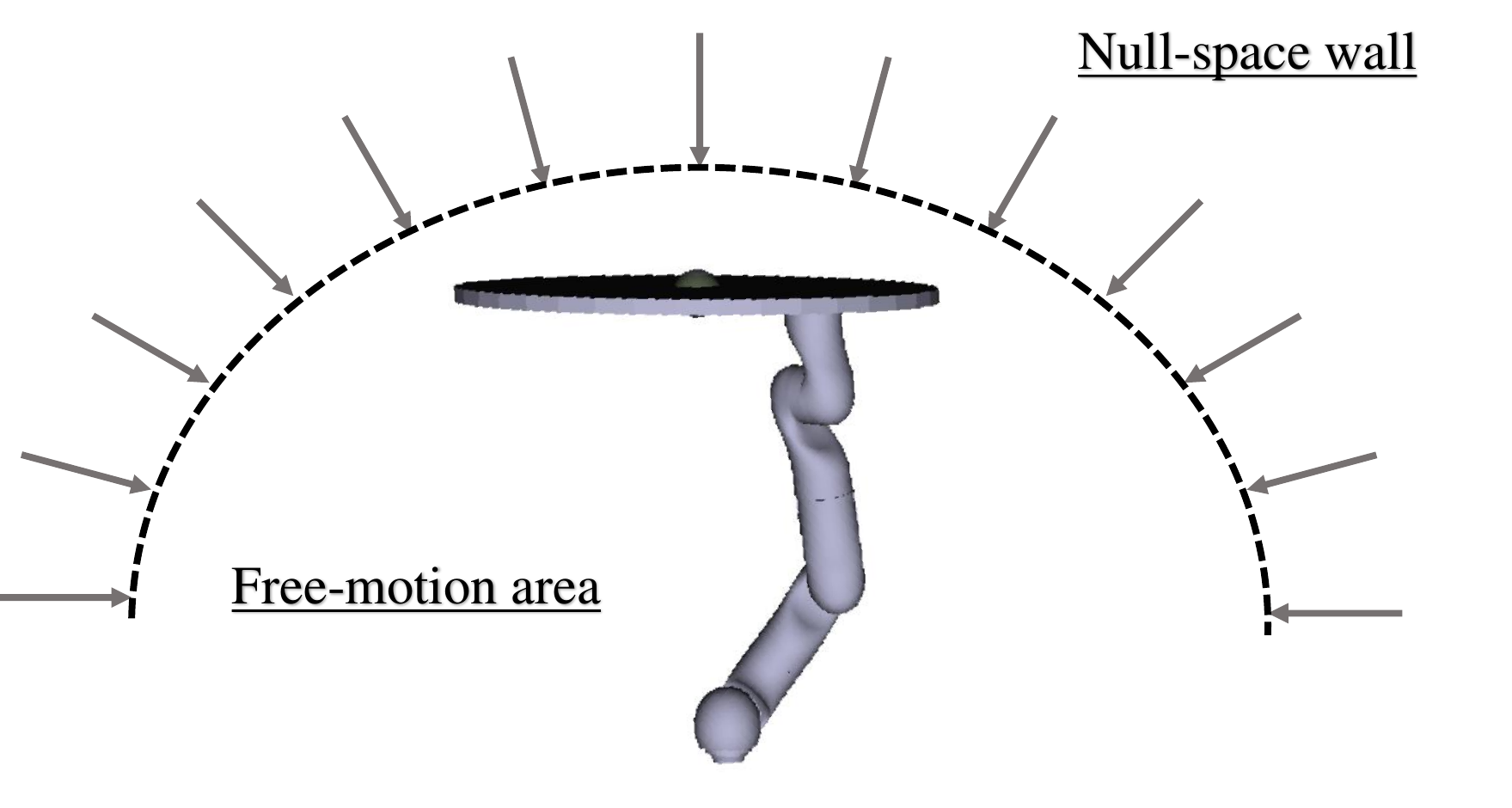}
 \caption{The Null-Space Wall concept.}
 \label{fig:ns_wall}
\end{figure}

On the other hand, if the base pose commanded by the operator is reachable without disturbing the end-effector, the operator will perceive lower forces, which are due to the imperfect tracking capabilities of the controller, as is usually the case for a P-F architecture in teleoperation. Therefore, for each end-effector pose there would be a region where the operator will perceive the robot in free-space motion and another where a wall penetration feeling will be perceived, which is the case when the commanded position is not feasible. An illustration of this behavior can be seen in Fig.~\ref{fig:ns_wall}.

\par
It is important to remark that, without the aforementioned haptic feedback the operator would have to rely on visual feedback in order to know if the flying base is following the commands. This could be a complex task since (1) the camera is usually attached to the base, which, therefore, does not appear in the frame, (2) it is hard to distinguish motion of the base from that of the end-effector in the camera images. On the other hand, the wall penetration feeling perceived by the operator would indicate that the desired direction of motion is not feasible.
\begin{figure}[b]
 \centering
 \includegraphics[trim={1cm 8cm 0.9cm 5.5cm},clip,width=\linewidth]{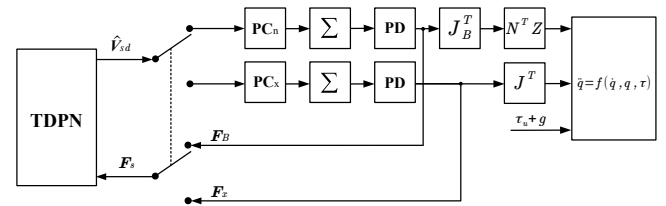}
 \caption{Block diagram of the proposed approach.}\label{fig:main_diag}
\end{figure}

\subsection{Bilateral Null-Space Teleoperation}
In order to allow bilateral teleoperation, the slave side of the P-F architecture (see Fig.~\ref{fig:drift_comp}) can be modified to implement a logic where the operator is able to switch between the primary task and the secondary one. For that purpose, the whole-body architecture depicted in Fig.~\ref{fig:main_diag} is proposed. In that architecture, the operator has the possibility of switching between the primary and the secondary task. 
\par
Initially, the initial values of $\mbi{g}_{x,des}$ and $\mbi{g}_{b,des}$ have to be set in the integrators represented by the summation blocks, which implement (\ref{eq:int_body}). When the switch is set for the primary task, the velocity commanded by the operator will be integrated into $\mbi{g}_{x,des}$ while $\mbi{g}_{b,des}$ will hold its value ($\mbi{V}_{x,des}(k)=\mbi{V}_{sd}(k)$, $\mbi{V}_{b,des}(k)=\mathbf{0}$). In addition, the force $\mbi{F_x}$ will be fed back through the communication channel while $\mbi{F}_b$ will act locally in order to keep the base at its last commanded pose. On the other hand, when the operator switches to the secondary task, the commanded velocity will be integrated into $\mbi{g}_{b,des}$ ($\mbi{V}_{b,des}(k)=\mbi{V}_{sd}(k)$, $\mbi{V}_{x,des}(k)=\mathbf{0}$) and the force $\mbi{F}_b$ will be sent back through the channel before going through the change of coordinates defined by $\Jb^{\mathit{\,T}}$ and the projection defined by $\mbi{N(q)}^{\mathit{T}}\mbi{Z(q)}$. In addition, $\mbi{\tau}_\mu + \g$ will be applied at all moments in order to compensate for gravity and diagonalize the matrix $\mu(\q,\dq)$ (see Section~\ref{sec:ctrl}).
\par
It is also important to note that, if the communication channel introduces time-delay and/or package loss, the teleoperation system is likely to become unstable. Therefore, in order to keep the communication channel passive, four PO-PC pairs can be implemented, two on the master and two on the slave side ($\mathbf{PC}_n$ and $\mathbf{PC}_x$ in Fig.~\ref{fig:main_diag}). These controllers will be responsible for monitoring and correcting the value of four potential functions, namely $W_{M,x}(k)$, $W_{S,x}(k)$, $W_{M,b}(k)$, and $W_{S,b}(k)$, which can be defined as
\begin{align} \label{eq:po}
&W_{M,x}(k) = E^S_{in,x}(k-T_b(k))-E^M_{out,x}(k)+E^M_{PC,x}(k-1) \, , \\
&W_{S,x}(k) = E^M_{in,x}(k-T_f(k))-E^S_{out,x}(k)+E^S_{PC,x}(k-1) \, , \\
&W_{M,b}(k) = E^S_{in,b}(k-T_b(k))-E^M_{out,b}(k)+E^M_{PC,b}(k-1) \, , \\
&W_{S,b}(k) = E^M_{in,b}(k-T_f(k))-E^S_{out,b}(k)+E^S_{PC,b}(k-1) \, .
\label{eq:po3}
\end{align} 
\par
The directions $in$ and $out$ of the energies are defined by the sign of the power variables $P^M(k)$ and  $P^S(k)$ \cite{ryu10}, which can be calculated as 
\begin{align}
    P^M(k)&= \mbi{V}_m(k)^{\mathit{T}}\hat{\mbi{F}}_m(k) \, , \\
    P^S(k)&= \hat{\mbi{V}}_{sd}(k)^{\mathit{T}}\mbi{F}_s(k) \, ,
\end{align}
where $\mbi{V}_m(k)$ and $\hat{\mbi{F}}_m(k)$ are the velocity of the master device and the force applied to it, as defined in Section~\ref{sec:tdpa}, while $\hat{\mbi{V}}_{sd}(k)$ and $\mbi{F}_s(k)$ will be the desired velocity and the computed force of the primary or the secondary task, depending on which one is being teleoperated (see Fig.~\ref{fig:main_diag}). Furthermore, the energy values can then be defined as
\begin{equation} \label{eq:EMx}
E^M_{x}(k)=
\begin{cases}
    E^M_{x}(k-1) + \Delta T P^M(k)           & \text{if } NS=0 \\
    E^M_{x}(k-1) & \text{else} \,,
\end{cases}
\end{equation}
\begin{equation} \label{eq:ESx}
E^S_{x}(k)=
\begin{cases}
   E^S_{x}(k-1) + \Delta T P^S(k)           & \text{if } NS=0 \\
   E^S_{x}(k-1) & \text{else} \,,
\end{cases}
\end{equation}
\begin{equation} \label{eq:EMb}
E^M_{b}(k)=
\begin{cases}
    E^M_{b}(k-1) + \Delta T P^M(k)           & \text{if } NS=1 \\
    E^M_{b}(k-1) & \text{else} \,,
\end{cases}
\end{equation}
\begin{equation} \label{eq:ESb}
E^S_{b}(k)=
\begin{cases}
   E^S_{b}(k-1) + \Delta T P^S(k)           & \text{if } NS=1 \\
   E^S_{b}(k-1) & \text{else} \,,
\end{cases}
\end{equation}
where $NS$ is the flag for null-space teleoperation. 
\begin{figure}[t]
 \centering
 \includegraphics[trim={4cm 0cm 4cm 0cm},clip,width=\linewidth]{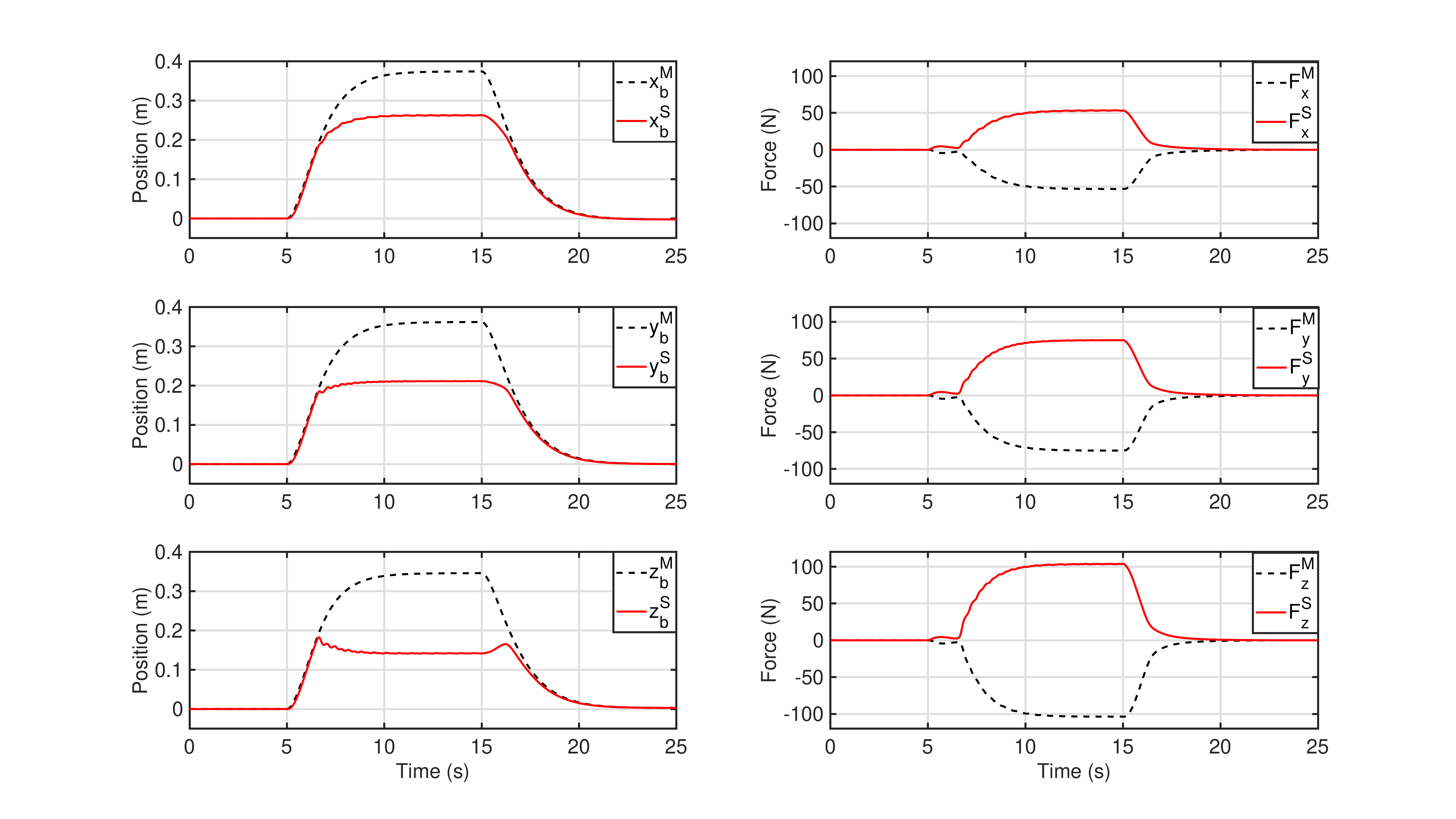}
 \caption{Null-space-wall penetration -- no time-delay.}
 \label{fig:haptics}
\end{figure}
\par
The use of separate energy functions and their dissipation only in the respective hierarchical level was chosen in order to prevent the loss of the energetic decoupling between them. However, the cross-coupled dissipation law presented in \cite{ott11} can also be applied without compromising the passivity of the system.

\section{Numerical Validation}
In order to validate the proposed whole-body teleoperation approach, three sets of numerical simulation were performed. In all of them, the master device was modeled as a non-redundant robot with dynamics as in (\ref{eq:dyn}) and the slave was the simulated SAM. Initially, the linear axes of the base were teleoperated in the null space of the end-effector, which was locally commanded to keep its initial pose. In that task whose purpose was to clearly show the behavior of the null-space wall, no time-delay was simulated. Fig.~\ref{fig:haptics} shows the positions and forces of the flying base, which shows that, between $t \approx 7$ and $t \approx 17$ seconds, the position commanded by the master $x_b^M$, $y_b^M$, and $z_b^M$ could not be followed by the slave because it reached the limits of the null-space motion. From that moment on, the interaction forces were increased as the master kept penetrating the null-space wall. The force in the opposite direction, which is fed to the master device, in this case, would inform the operator that that the commanded position is not reachable. 
\begin{figure}[t]
 \centering
 \includegraphics[trim={4cm 0cm 4cm 0cm},clip,width=\linewidth]{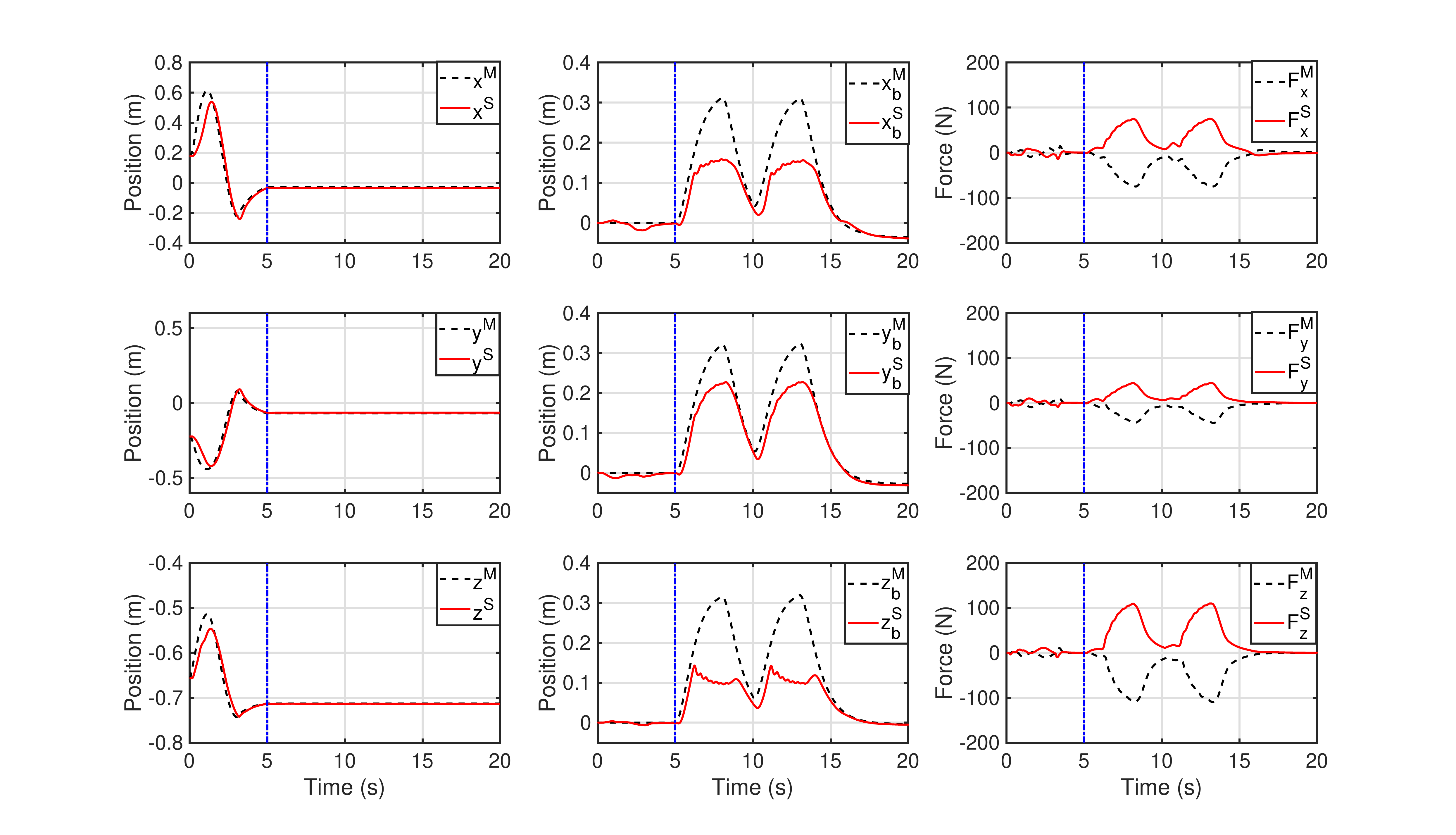}
 \caption{Whole-body teleoperation, positions for primary (left) and secondary task (middle), and haptic forces (right) -- $T_{rt}=300$~ms.}
 \label{fig:decoupled1}
 \includegraphics[trim={4cm 1cm 4cm 0cm},clip,width=\linewidth]{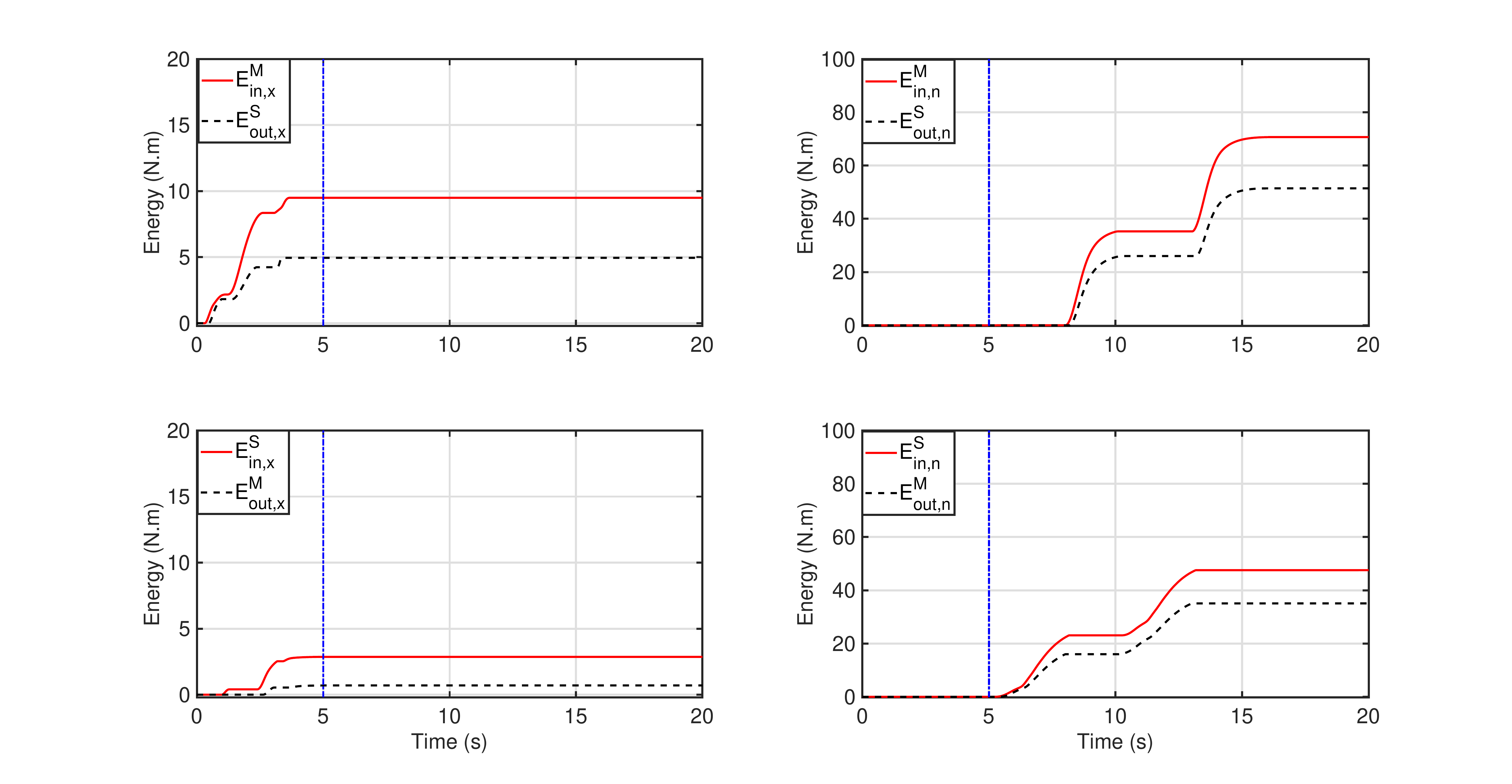}
 \caption{Whole-body teleoperation, input and output energies for primary (left) and secondary task (right)  -- $T_{rt}=300$~ms. }
  \label{fig:decoupled2}
\end{figure}
\begin{figure}[t]
 \centering
  \includegraphics[trim={4cm 0cm 4cm 0cm},clip,width=\linewidth]{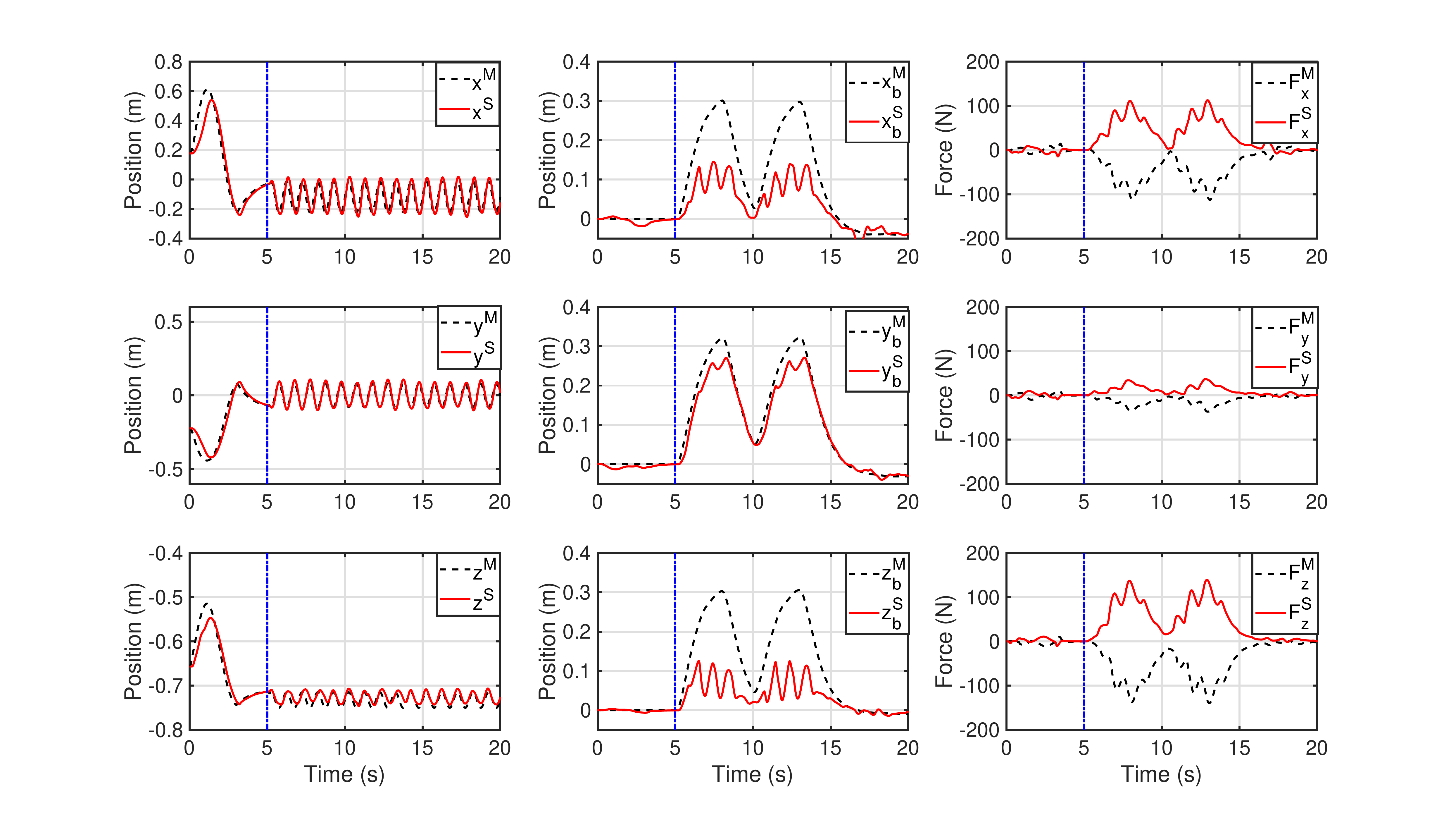}
 \caption{Whole-body teleoperation with autonomous end-effector motion during null-space teleoperation,  positions for primary (left) and secondary task (middle), and haptic forces (right) -- $T_{rt}=300$~ms.}
 \label{fig:coupled1}
 \includegraphics[trim={4cm 0cm 4cm 0cm},clip,width=\linewidth]{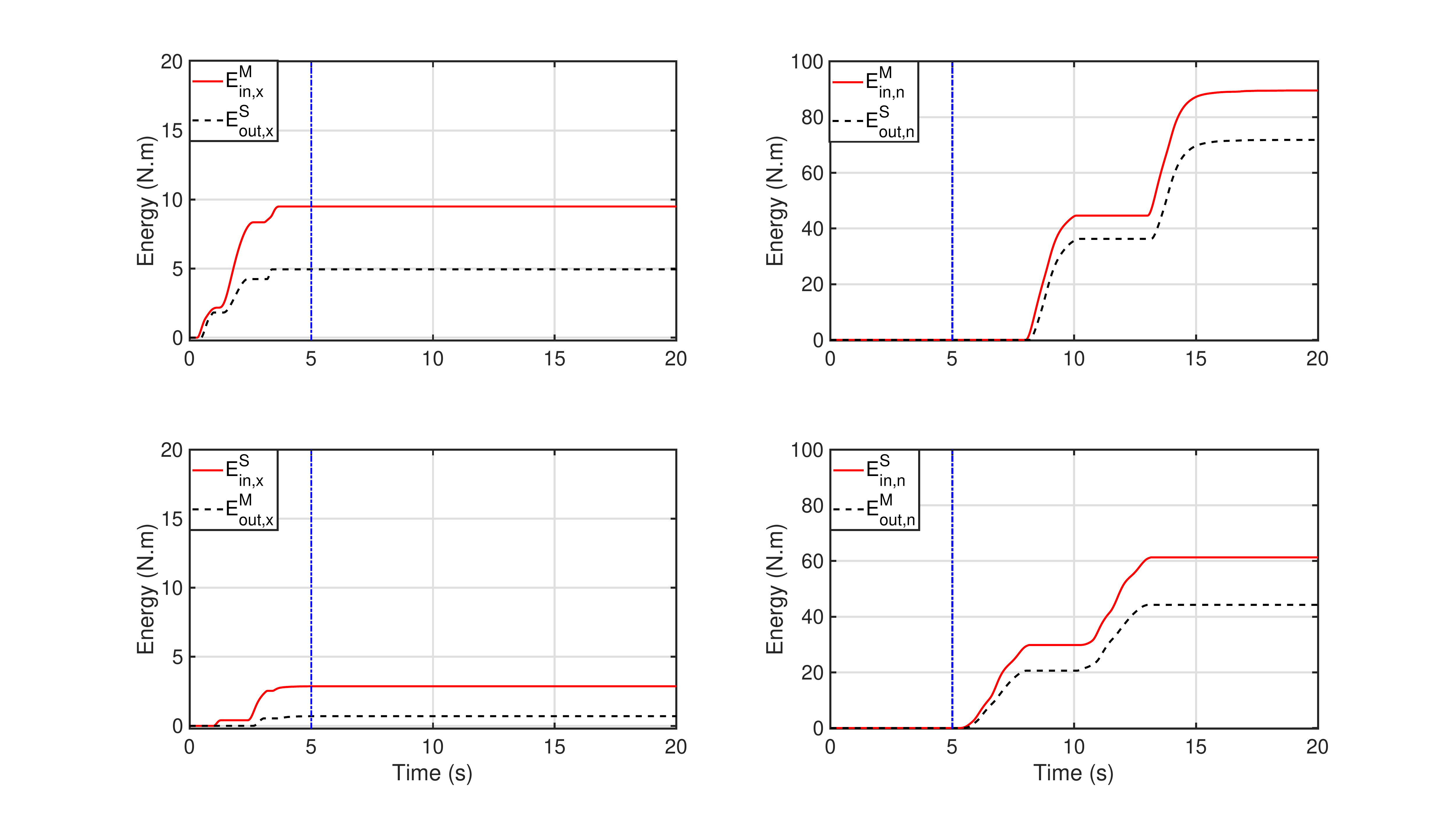}
 \caption{Whole-body teleoperation with autonomous end-effector motion during null-space teleoperation, input and output energies for primary (left) and secondary task (right) -- $T_{rt}=300$~ms.}
  \label{fig:coupled2}
\end{figure}
\par
The second task involved teleoperating the end-effector under 300 ms round-trip delays and, at $t = 5$ seconds (dashed vertical line), switching to the base teleoperation. For the sake of simplicity, the weighting matrices $\mbi{\Gamma}(k)$ and $\mbi{\Psi}(k)$ described in Section~\ref{sec:tdpa} were set to be identity matrices. Additionally, the drift compensator from \cite{coelho19} was applied in order to avoid drift caused by TDPA. Fig.~\ref{fig:decoupled1} displays the master and slave positions of both the end-effector and the flying base, and the force transmitted through the communication channel, which before 5 seconds is relative to the free motion of the end-effector and, after that, is due to the motion of the flying base, which hits the null-space wall twice. It can be noted that the end-effector motion also required the flying base to slightly move, which is due to its higher level of priority in the control structure. Adding to that, since the end-effector was teleoperated in free-space motion, the force feedback provided to the user in that period is much smaller that the one generated by the reaching of the null-space limits, which resembles wall-contact forces. It can also be seen that, since there is a dependency between the wall position in all three axes, the wall is not completely flat. However, since the end-effector is steady, the curvature of the wall is reasonably smooth. Fig.~\ref{fig:decoupled2} presents the two energy flow of the channel. It can be noted that the TDPA was able to keep the channel passive at all time steps.
\par
The last task aimed at analyzing the behavior of the wall when the end-effector is moving. For that purpose, after switching to null-space teleoperation, the end-effector was commanded to move autonomously in open loop. It can be seen from Fig.~\ref{fig:coupled1} that the limits of the null-space wall changed according to the end-effector motion, making the slave oscillate when in contact with the wall. It is also important to remark that the haptic forces displayed a high frequency sinusoidal wave, which is due to the end-effector motion. This feature could also be beneficial to inform the operator in case the end-effector is externally disturbed during teleoperation of the flying base. In can be noted that, despite not being included in the theoretical analysis, a moving null-space wall seems not to destabilize the system.

\section{Conclusion and Future Work}
\label{sec:conc}
This paper presented a framework for performing stable bilateral teleoperation of both the end-effector and the base of redundant aerial manipulators. This feature is especially beneficial when a video camera is attached to the flying base whose position has to be changed without disturbing the end-effector. It has been shown that applying a hierarchically-decoupling controller and feeding back the secondary control force before being multiplied by the projection term provides the operator with a wall-contact feeling whenever the base is commanded to move to an unreachable position. This characteristic was confirmed by numerical simulation data, which also showed that the proposed framework can render the system passive in the presence of communication delays. Future work will involve applying the proposed framework to teleoperation of arbitrary hierarchy levels \cite{ott15}. In addition, an extension to a four-channel architecture \cite{artigas16} to include measured-force feedback will be studied.
\FloatBarrier
\bibliographystyle{IEEEtran}
\bibliography{root}

\end{document}